\documentclass[letterpaper, 10 pt, conference]{ieeeconf}  

\IEEEoverridecommandlockouts                              

\usepackage{graphics} 
\usepackage{epsfig} 
\usepackage{mathptmx} 
\usepackage{times} 
\usepackage{amsmath} 
\usepackage{amssymb}  
\usepackage[dvipsnames]{xcolor}
\usepackage{dblfloatfix} 

\title{\LARGE \bf
A Feasible Workflow for Retinal Vein Cannulation \\
in  \textit{Ex Vivo} Porcine Eyes with Robotic Assistance
}

\author{Peiyao Zhang$^{1}$, Peter Gehlbach$^{2}$, Marin Kobilarov$^{1}$, and Iulian Iordachita$^{1}$
\thanks{$^{1}$Peiyao Zhang, Marin Kobilarov, and Iulian Iordachita are with the Department of Mechanical Engineering and the Laboratory for Computational Sensing and Robotics (LCSR), Johns Hopkins University, Baltimore, MD 21211, USA 
        {\tt\small \{pzhang24, mkobila1, iordachita\}@jhu.edu}}%
\thanks{$^{2}$Peter Gehlbach is with the Wilmer Eye Institute, Johns Hopkins University, Baltimore, MD 21211, USA 
        {\tt\small pgelbach@jhmi.edu}}%
}

\begin{document}
\maketitle
\thispagestyle{empty}
\pagestyle{empty}

\begin{abstract}
A potential Retinal Vein Occlusion (RVO) treatment involves Retinal Vein Cannulation (RVC), which requires the surgeon to insert a microneedle into the affected retinal vein and administer a clot-dissolving drug. This procedure presents significant challenges due to human physiological limitations, such as hand tremors, prolonged tool-holding periods, and constraints in depth perception using a microscope. This study proposes a robot-assisted workflow for RVC to overcome these limitations. The test robot is operated through a keyboard. An intraoperative Optical Coherence Tomography (iOCT) system is used to verify successful venous puncture before infusion. The workflow is validated using 12 \textit{ex vivo} porcine eyes. These early results demonstrate a successful rate of 10 out of 12 cannulations (83.33\%), affirming the feasibility of the proposed workflow.
\end{abstract}

\section{INTRODUCTION}
RVO is a vascular disorder resulting from the blockage of one or more retinal veins responsible for carrying away the blood from the retina. It ranks as the second leading cause of retinal vascular blindness in the Western World, followed by diabetic retinopathy. Untreated RVO can result in vision loss by macular edema, ocular neovascularization, hemorrhage, retinal detachment, and others \cite{Ip}. There are two main types of RVO - Central Retinal Vein Occlusion (CRVO) and Branch Retinal Vein Occlusion (BRVO). According to research by Rogers et al. \cite{Rogers}, RVO affects approximately 16.4 million adults worldwide. Among these people, 2.5 million are affected by CRVO, while 13.9 million are affected by BRVO. BRVO is more prevalent than CRVO but poses more significant challenges for robotic treatment due to the smaller caliber of branch veins. The mean central retinal vein equivalent diameter is reported to be 251.0$\mu m$ \cite{Drobnjak}, while location-dependent, the mean branch retinal vein size is measured to be 151.32$\mu m$ \cite{Goldenberg}.

The current approach to treating RVO primarily focuses on alleviating its associated complications. Standard methods include grid or panretinal laser photocoagulation, intravitreal anti-Vascular Endothelial Growth Factor (anti-VEGF) drugs, and intravitreal steroids \cite{Berger}. However, these techniques do not address the root cause of the vein obstruction. In contrast, RVC is an emerging experimental method with the potential for dissolving blood clots and restoring blood flow to the retina \cite{Willekens}.
\setcounter{figure}{0} 
\begin{figure}[thpb]
  \centering
  \includegraphics[width=\columnwidth]{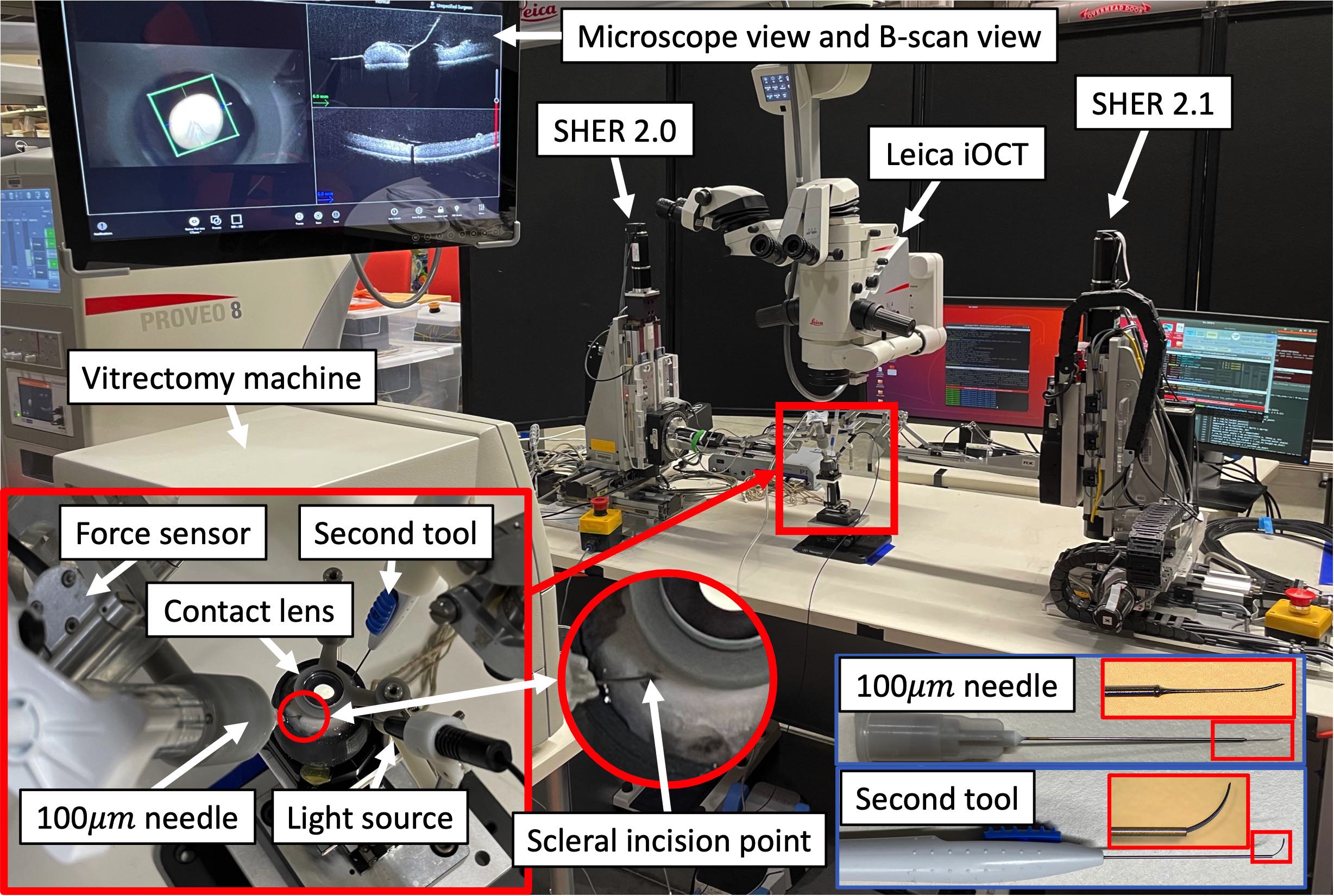}
  \vspace{-18pt}
  \caption{Experimental setup: two SHERs control the needle and the medical spatula; Leica iOCT provides a top-down microscope view and B-scans; the vitrectomy machine provides light source and infusion pressure; the force sensor measures the handle force.}
  \label{fig:experimental_setup}
  \vspace{-20pt}
\end{figure}
This procedure involves inserting a needle into the occluded vein to deliver anticoagulating drugs, such as recombinant tissue Plasminogen Activator (rtPA) \cite{van Overdam} and ocriplasmin \cite{de Smet}. It has been validated using various eye models, including eye phantoms \cite{Gerber}, \cite{Gonenc}, open-sky porcine eyes \cite{Kadonosono}, \textit{in vivo} porcine eyes \cite{Willekens}, and human patients \cite{van Overdam}, \cite{Gijbels}. Due to the tiny vein size, \textit{in vivo} experiments often use glass micropipettes for cannulation \cite{Willekens}, \cite{van Overdam}, \cite{de Smet}, as they can be made thinner and sharper than metal needles. However, glass micropipettes are exceedingly fragile and are challenging to visualize under a microscope, leading to reported instances of broken needles \cite{Willekens}, \cite{de Smet}, \cite{Kadonosono}, \cite{Koen}. While metal needles are more robust and do not require tip coatings to enhance the visibility, there is a very high cost (hundreds of US dollars) for metal needles with a tip size of 60$\mu m$ or less.
\setcounter{figure}{2} 
\begin{figure*}[!b]
  \centering
  \includegraphics[scale=0.155]{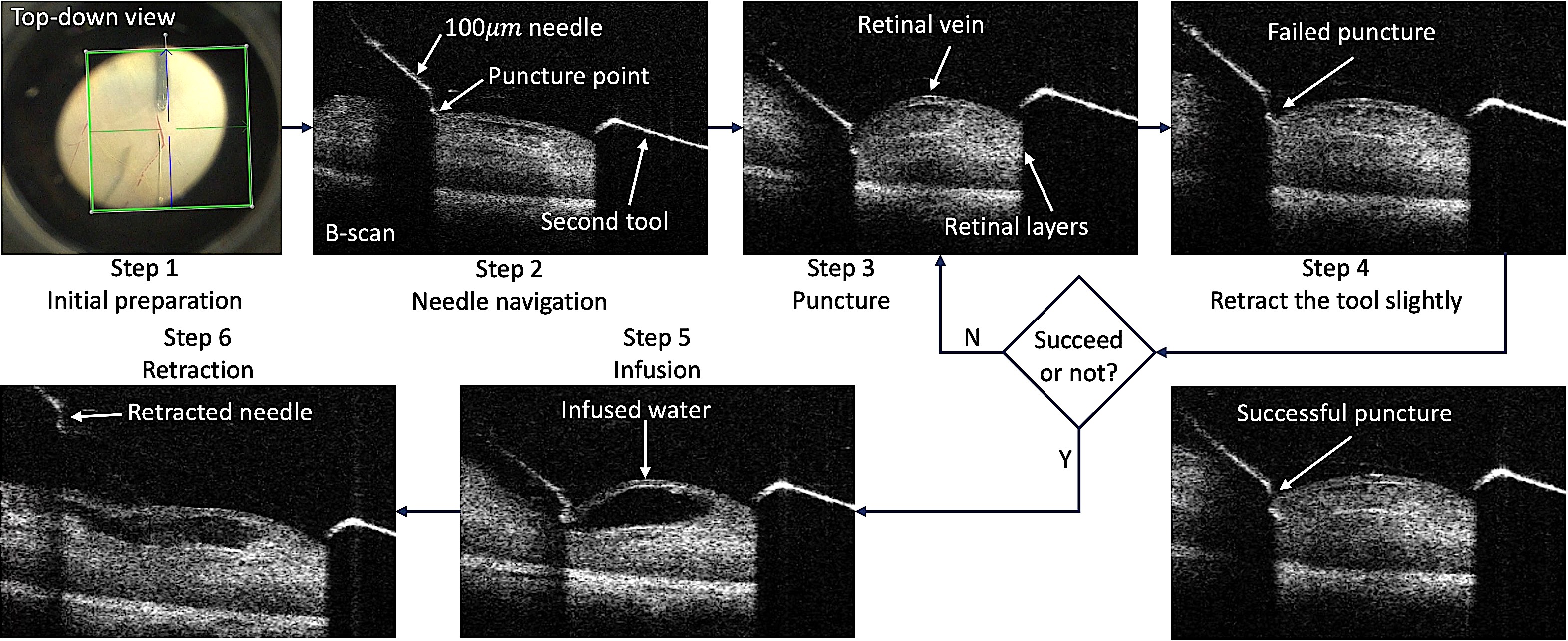}
  \vspace{-8pt}
  \caption{Workflow of proposed robot-assisted RVC method.}
  \label{fig:workflow}
\end{figure*}

Most successful RVC experiments rely on robotic assistance due to human physiological limitations. The reported hand tremor in humans is on the order of 180$\mu m$ \cite{Riviere}. Given the mean diameter of the retinal vein, consistently completing RVC without assistance is very challenging for humans. Various robotic systems have been developed to mitigate hand tremors, including hand-held \cite{Yang}, teleoperated \cite{Edwards}, and co-manipulated systems \cite{Ali}. Additionally, the surgeon's hand must remain static during the drug infusion phase, posing a challenge for the human operator to hold the tip still through prolonged infusion periods.
\setcounter{figure}{1} 
\begin{figure}[thpb]
  \centering
  \includegraphics[width=\columnwidth]{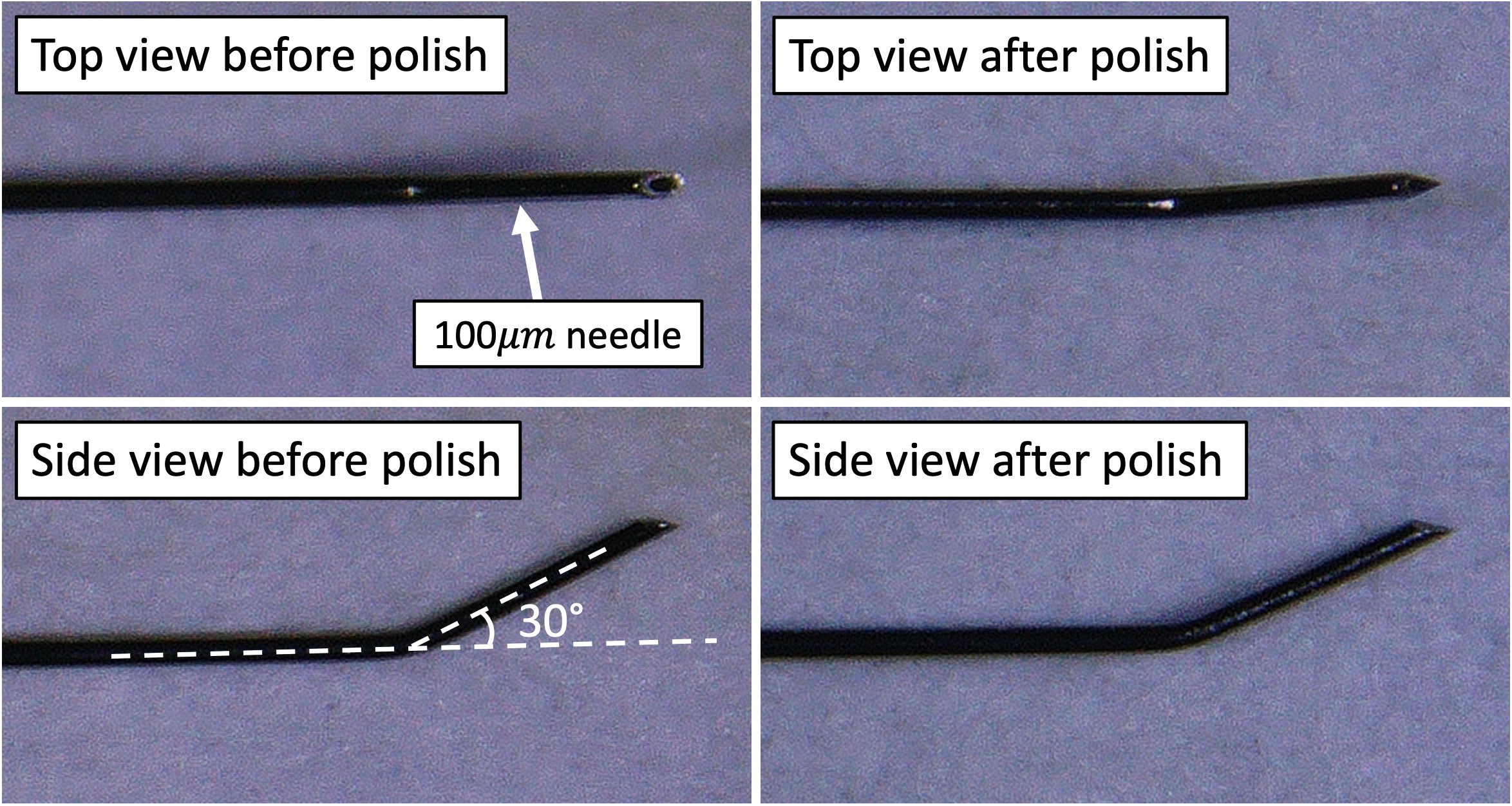}
  \vspace{-16pt}
  \caption{Two views of the needle tip before and after needle polish.}
  \label{fig:polish}
  \vspace{-18pt}
\end{figure}
The second limitation arises from limits on human depth perception for surgeons relying on the single top-down microscope view. The shadow of the tool is often used to assess the depth \cite{Zhang}. However, it can be challenging to determine whether the needle is inside the vessel lumen when using only this method. A further limitation involves maintaining the Remote Center of Motion (RCM) constraint at the scleral incision point, Fig. \ref{fig:experimental_setup}. Large forces applied to this point should be avoided to prevent eye rotation, striae to the cornea, or damage to the eye. Although maintaining the RCM constraint may be demanding for surgeons, robots can achieve it mechanically \cite{Gerber} or through using algorithmic control \cite{Zhang}.

As the main contribution, in this study, we present a feasible workflow for RVC using 100$\mu m$ beveled metal needle in \textit{ex vivo} porcine eyes. Two Steady Hand Eye Robots \cite{Ali} are deployed for this purpose. The main robot (SHER 2.0) holds the needle, while the auxiliary robot (SHER 2.1) holds a medical spatula. We use the keyboard to control the main robot's movement from the initial position to the retraction of the needle. The robot enforces the RCM constraint throughout the workflow. A cross-sectional image (B-scan) of the vein is obtained using an iOCT system to verify successful puncture into the vessel lumen before infusion. An OCT system is a medical imaging technology that uses near-infrared light waves to capture cross-sectional images of biological tissues. An iOCT system integrates an OCT system into the microscope, providing real-time imaging during a surgical procedure. Ten successes out of twelve attempts at RVC are accomplished using this workflow.

\section{METHOD}
\subsection{Experimental Setup}
Fig. \ref{fig:experimental_setup} shows our experimental setup, including two SHERs (Johns Hopkins University), an iOCT system (Leica Proveo 8), a light source (Alcon Accurus 20G Sapphire Wide Angle Endoilluminator), a contact lens (Oculus Surgical Super View Hassan-Tornambe Contact Lens), a 100$\mu m$ needle (MedOne MicroTip Beveled Cannula 25g/40g), a spatula (Dorc extendible curved spatula), and a 3D-printed eye socket to hold the eye. Both SHERs feature 5 degrees of freedom, consisting of 3 translational joints and 2 rotational joints \cite{Ali}. The Leica iOCT provides a top-down microscope view and B-scan images of the region of interest. The contact lens offers a 36-degree field of view for a clear view of the retinal target area. The needle is affixed to a syringe connected to the Viscous Fluid Control (VFC) mode of the Alcon Accurus vitrectomy machine. This allows us to regulate the maximum infusion pressure, facilitating a steady and stable infusion process. All the components are used previously in the operating room during animal trials. The experiments are conducted by an expert user of the SHER, the Leica iOCT system, and the vitrectomy machine.

\subsection{Initial Preparation}
\label{initial_preparation}
Porcine eyes were harvested and obtained from a local butcher shop on the day of harvest. The surrounding tissue was removed using a forceps (STORZ Castroviejo Suturing Forceps E1796) and scissors (STORZ Westcott Type Utility Scissors E3322). The needle underwent additional sharpening with a polisher (Buehler Metaserv 2000), as shown in Fig. \ref{fig:polish}. Three pars plana sclerotomies were made for the needle, spatula, and light source insertions using trocars (Alcon 20G Trocar). There are some inherent limitations when working with cadaveric porcine eyes. Firstly, performing a complete vitrectomy is unnecessary, as we injected saline just above the RVC position to simulate the absence of vitreous gel. The resulting bleb eliminates vitreous compression as the needle approaches the vein. The second limitation stems from the lack of blood pressure and an occluding thrombotic obstruction in the veins of cadaveric porcine eyes. Puncturing the vein is more challenging without blood pressure due to the collapse of the blood vessel walls and lack of a lumen.
\begin{table}
\caption{Movement Keys of Keyboard Controller}
\vspace{-12pt}
\label{table:keys}
\begin{center}
\begin{tabular}{|c|c|}
\hline
\multicolumn{1}{|p{2cm}|}{\centering Keys}
& \multicolumn{1}{p{5cm}|}{\centering Movement Description}
\\
\hline
Left arrow & Moving along negative X-axis\\
\hline
Right arrow & Moving along positive X-axis\\
\hline
Up arrow & Moving along positive Y-axis\\
\hline
Down arrow & Moving along negative Y-axis\\
\hline
D & Moving along negative Z-axis\\
\hline
U & Moving along positive Z-axis\\
\hline
P & Quick push for puncture\\
\hline
R & Retracting the needle\\
\hline
\end{tabular}
\end{center}
\vspace{-12pt}
\end{table}
\begin{table}
\caption{Mean Time Duration}
\vspace{-12pt}
\label{table:mean_time}
\begin{center}
\begin{tabular}{|c|c|}
\hline
\multicolumn{1}{|p{5cm}|}{\centering Step}
& \multicolumn{1}{p{1.5cm}|}{\centering Mean time}
\\
\hline
Step 2 Needle navigation & 57.45s\\
\hline
Step 3 puncture and step 4 retract slightly & 43.55s\\
\hline
Step 5 infusion & 67.58s\\
\hline
Step 6 retraction & 19.99s\\
\hline
\end{tabular}
\end{center}
\vspace{-20pt}
\end{table}To overcome this, the spatula is positioned close to the cannulation target. By compressing one side of the blood vessel, we locally flush the retained blood toward the puncture site, simulating blood pressure. Finally, the needle is inserted into the eye and guided to the target. The RCM point is recorded when the needle tip aligns with the scleral incision point. The needle tip position is derived through robot kinematics \cite{Ali}, \cite{Murray}.

\subsection{Workflow}
The six-step workflow is shown in Fig. \ref{fig:workflow}. The initial preparation is discussed in Section \ref{initial_preparation}. The needle navigation step involves the user controlling the robot to reach the desired puncture position. This is accomplished through two substeps: navigating the needle using the top-down camera view to align the needle tip with the target and moving the needle downward until it makes contact with the superficial vessel wall. IOCT verifies this contact. The puncture step is executed as a quick push, which is an effective method to overcome the resistance of the blood vessel. A slow needle insertion deforms and flattens the vessel with needle entry. Following needle advancement, the needle is retracted slightly to verify whether a successful puncture into the vessel lumen has occurred. If not, steps 3 and 4 are repeated until a successful puncture is verified. A successful puncture is determined when the needle tip is confirmed to be intraluminal. This is confirmed using the VFC mode of the vitrectomy machine to perform a water flush. After injection, the needle is retracted.

\setcounter{figure}{3}
\begin{figure}[thpb]
  \centering
  \includegraphics[width=\columnwidth]{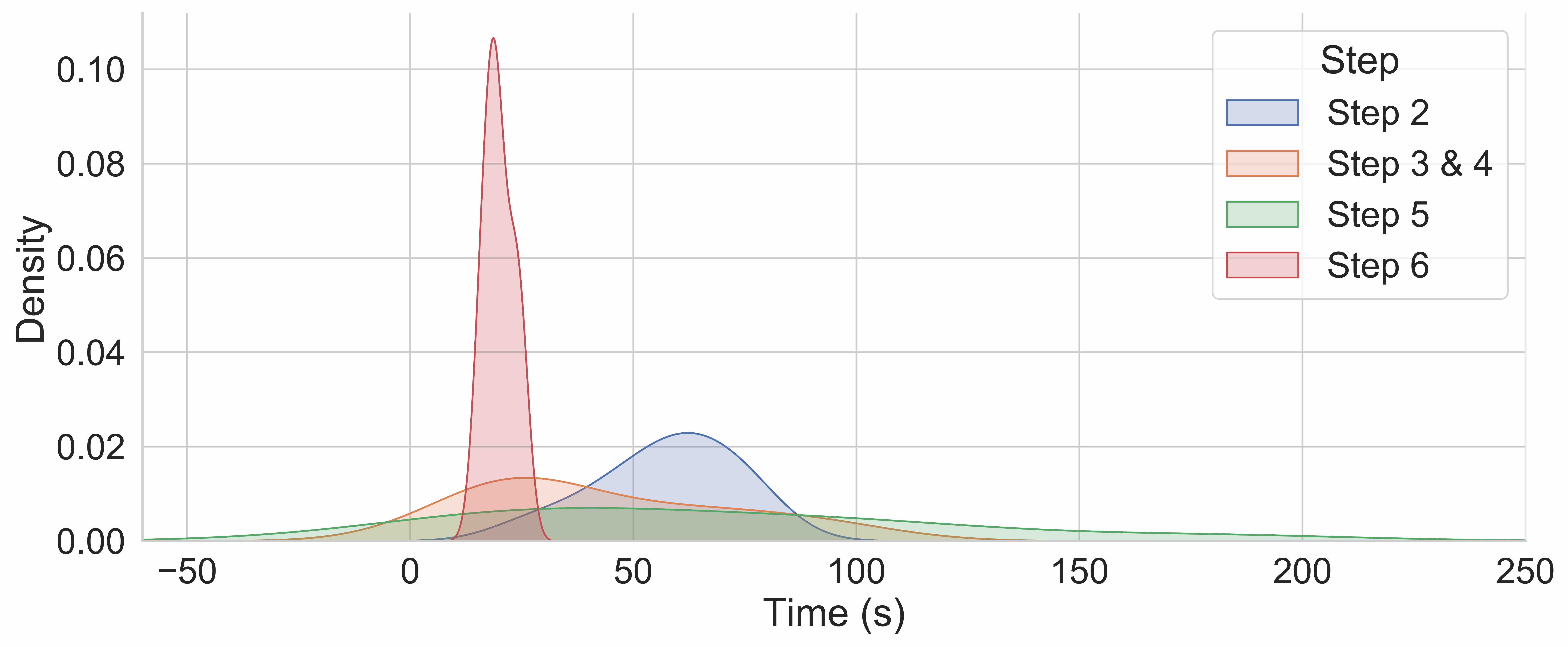}
  \vspace{-22pt}
  \caption{Time distributions for each step.}
  \label{fig:distributions}
  \vspace{-8pt}
\end{figure}
\setcounter{figure}{4}
\begin{figure}[thpb]
  \centering
  \includegraphics[width=\columnwidth]{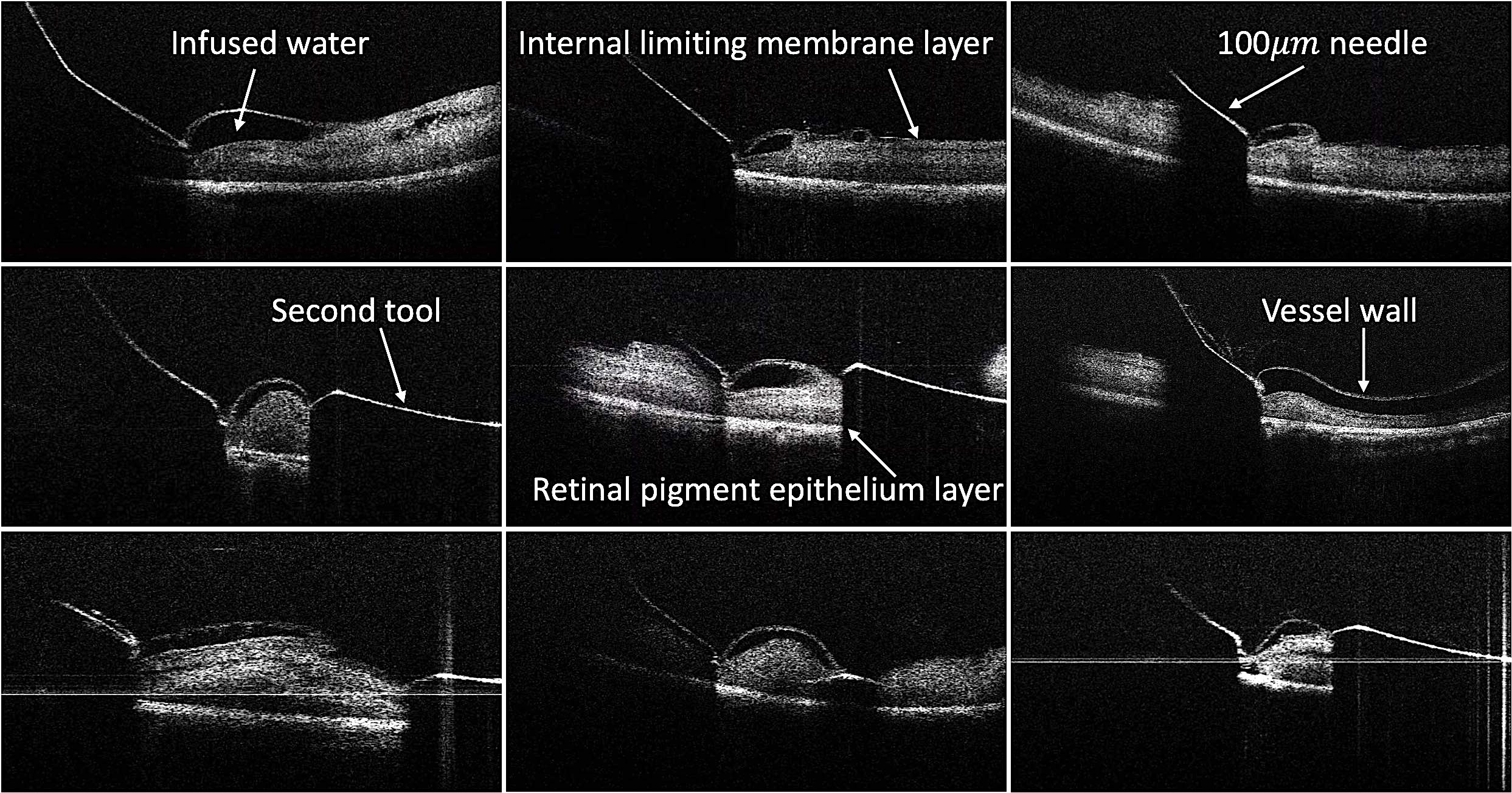}
  \vspace{-18pt}
  \caption{B-scans of targeted blood vessels after infusion.}
  \label{fig:infusion_results}
  \vspace{-16pt}
\end{figure}

\subsection{Keyboard Controller} 
The keyboard controller has eight basic movement keys as illustrated in Table \ref{table:keys}. The linear velocities for all movements are set at 0.2$mm/s$, except for the puncture, which is configured at 5.4$mm/s$. To enforce the RCM constraint, the angular velocity of the tool shaft is adjusted to minimize the error between the current tool shaft orientation and the line passing through both the RCM point and the tool tip. As mentioned, the RCM point is recorded when the needle is inserted into the eye. Suppose the current tool rotation matrix is denoted as $R_c$ and the desired rotation matrix as $R_d$, the rotation error is defined using $SO(3)$ logarithm function as $\log(R_c^TR_d)$. This skew-symmetric matrix is then converted into a vector. If the norm of this error vector is less than 0.1 degrees, the rotation of the needle is stopped. When a key is pressed, the robot initiates the corresponding needle movement in real time. If the key is released, the robot stops instantaneously.

\section{EXPERIMENTS AND RESULTS}
Our workflow was tested on 12 \textit{ex vivo} porcine eyes, resulting in 10 successful infusions and two failure cases. In one instance, the failure occurred due to residual air inside the needle, leading to the formation of air bubbles during the infusion step, thereby blocking the view. The second failure was caused by insufficient intraluminal blood to allow puncture. Each of these causes of failure would not be present \textit{in vivo}. The mean time duration for ten successful trials was recorded for steps 2 to 6, as shown in Table \ref{table:mean_time}. The time distributions for each step are shown in Fig. \ref{fig:distributions}. Although the initial preparation time was not recorded, it took approximately 25 minutes for each eye. The infusion pressure was set at 12$mmHg$. B-scans of the targeted blood vessel after infusion were presented in Fig. \ref{fig:infusion_results}. Clear empty vessels were created as the infused water flushed away the blood. In a specific trial, Fig. \ref{fig:velocity_and_force} showed needle tip velocities derived from robot kinematics and corresponding handle forces measured by a NANO 17 force/torque sensor from ATI at the robot's end-effector, Fig. \ref{fig:experimental_setup}.
\setcounter{figure}{5}\begin{figure}[thpb]
  \centering
  \includegraphics[width=\columnwidth]{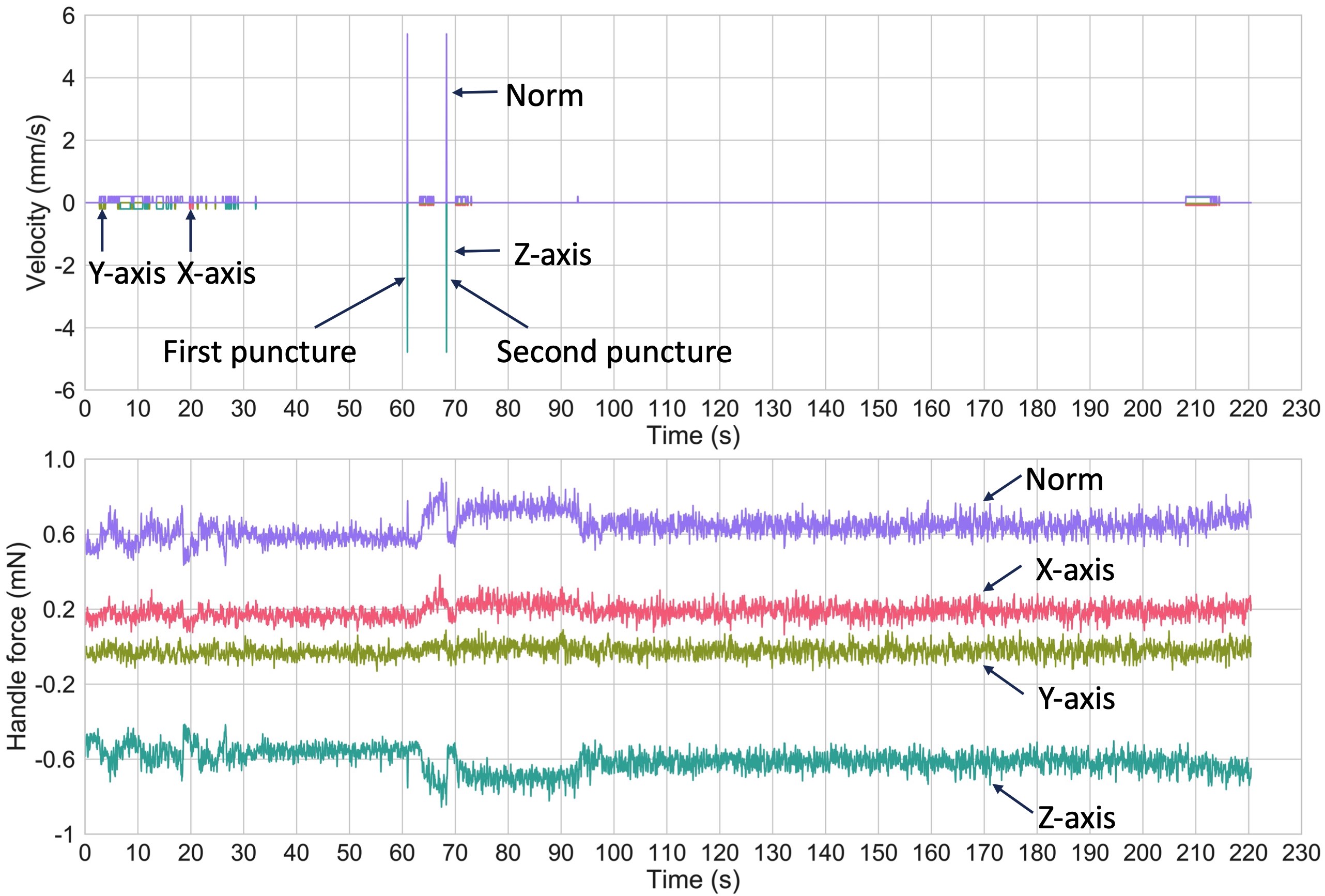}
  \vspace{-22pt}
  \caption{Velocities and handle forces along the XYZ-axes and their corresponding norms for a trial.}
  \label{fig:velocity_and_force}
  \vspace{-6pt}
\end{figure}
\setcounter{figure}{6}
\begin{figure}[thpb]
  \centering
  \includegraphics[width=\columnwidth]{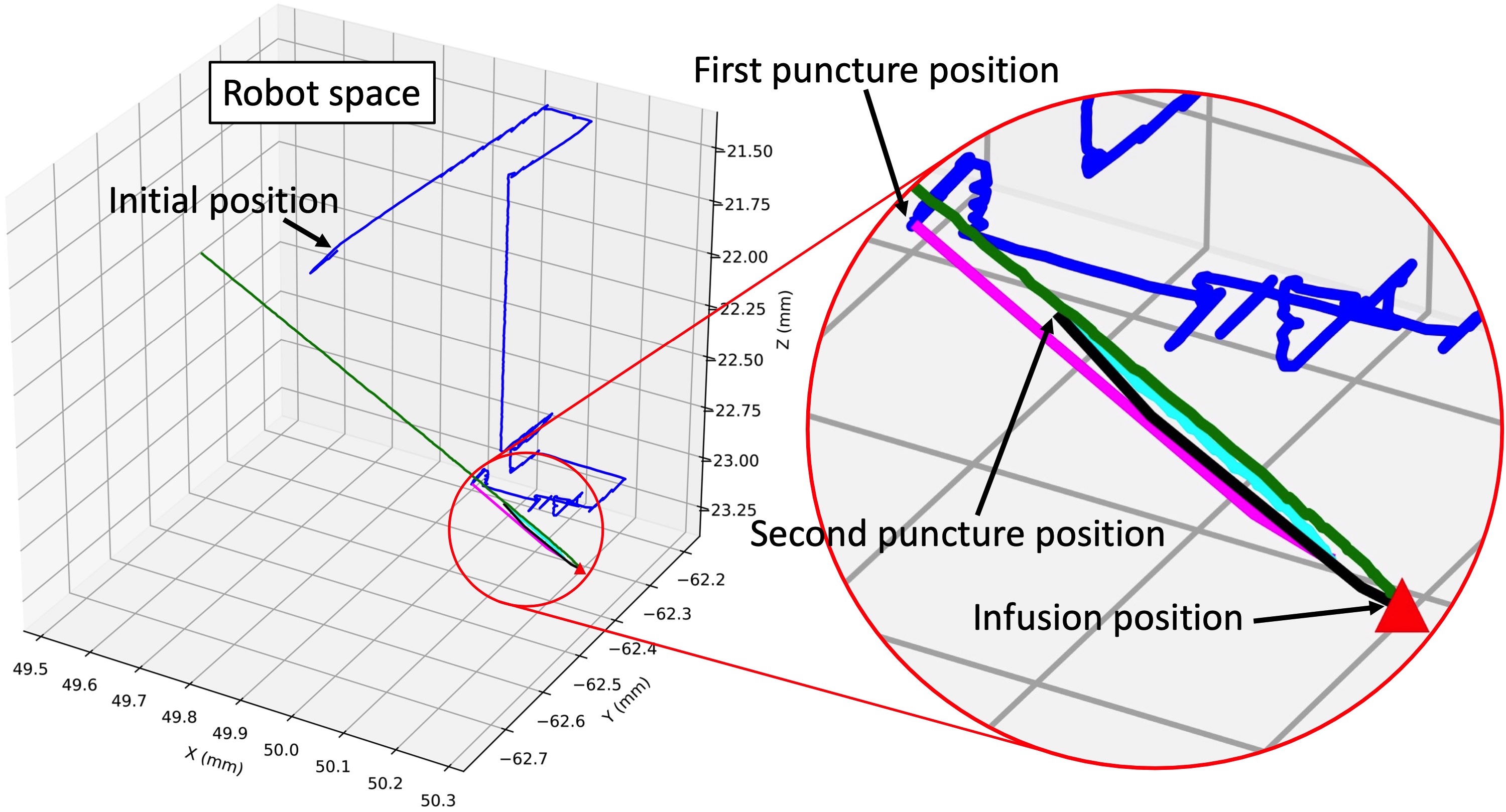}
  \vspace{-20pt}
  \caption{Trajectory of the needle tip during a trial. (Blue) Step 2: needle navigation; (Magenta and black) Step 3: puncture; (Cyan) Step 4: retract the tool slightly; (Red triangle) Step 5: infuison; (Green) Step 6: retraction.}
  \label{fig:trajectory}
  \vspace{-16pt}
\end{figure}Notably, this measured force was not at the scleral insertion point or the contact force between the needle tip and the tissue. Nevertheless, discernible force changes were observed at the two punctures. Fig. \ref{fig:trajectory} demonstrates the plotted needle tip trajectory in robot space for this trial, with the first insertion distance was recorded at 335$\mu m$ and the second insertion distance was 287$\mu m$. The mean accuracy of the robot's translational motion for each axis is measured to be 5$\mu m$ \cite{Shervin}.



\section{DISCUSSION}
A comparatively high success rate of 83.33\% indicates the feasibility of our workflow for RVC using a metal needle. The failure cases show the importance of thoroughly examining the needle before insertion to eliminate any potential air inside. Additionally, fresh tissue is essential to minimize vessel collapse. Cornea opacity in cadaveric porcine eyes reduces the image quality in the microscope and iOCT-scan. Various factors contribute to each step's duration, including the needle's initial position, eye freshness, and the operator's familiarity with the robot and the controller. Here, the infusion time differs from human studies, which could last around 6 minutes \cite{Koen}. A reason is that the needle used in this study is large, and our goal is to demonstrate the feasibility of our workflow for successful cannulation rather than prolonged infusion. The puncture velocity in this study is specifically chosen for cadaveric porcine eyes. While multiple velocities are tested, we believe the current puncture velocity yields the best results in this setting. However, we acknowledge that puncture velocity may vary based on the blood pressure level of the vein. Lower blood pressure may necessitate a higher velocity for successful puncture, possibly explaining the need for multiple attempts in some trials. In Fig. \ref{fig:trajectory}, the second insertion distance is shorter than the first due to the user needing to press and release the key immediately to control puncture movement. Consistency in key press time is difficult for the user. In future work, we aim to replicate this workflow using machine learning methods to automate some or all steps. This includes setting puncture time as a fixed value and determining the needle retraction distance in step 4 based on the insertion distance. 

This study has some limitations. The physical size of the metal needle makes it nearly impossible to catheterize the vessel lumen fully. Successful puncture confirmation is based on identifying a portion of the needle tip inside the vessel lumen sufficient to allow successful fluid infusion. This implies that the drug may also flow into the vitreous chamber during infusion. A potential solution is complete cannulation of the vessel with smaller needles. Another limitation is the absence of a post-analysis for the puncture point. Blood reflux is observed if excessive water is infused into the vessel lumen. Despite the beveled needle used in this work, the puncture orientation aligns with the main tool shaft. Surgeons might find it easier to perform RVC if the cannulation is tangential to the retinal vein \cite{van Overdam}. However, achieving punctures using this technique in this study is hindered by the resilience of the retinal vein. The needle tip slides along the superficial vessel wall, causing corresponding deformations. Currently, the B-scan position is manually controlled. It may not align with the needle tip after the puncture, necessitating further adjustment of the scanning position, which is time-consuming. A possible future solution is using a volume scan to extract the desired B-scan. Additionally, we assume that the eye remains stationary throughout the workflow, which might not reflect real surgical conditions. However, it has been shown that this problem can be solved by using optical flow methods to track the eye movement \cite{Zhang}.

\section{CONCLUSIONS}
This study presents a feasible workflow for robot-assisted RVC using a 100$\mu m$ metal needle. A keyboard controller is implemented to navigate the needle and perform the puncture. The workflow is validated using 12 \textit{ex vivo} porcine eyes, resulting in 10 successful cannulations. The puncture and cannulation are verified through an iOCT B-scan of the puncture site. The mean time duration for the entire workflow, excluding the initial preparation, is approximately 3 minutes, which is reasonable. The current study is the first step toward clinical trials. Our future work aims to automate some or all steps using deep learning methods in \textit{in vivo} experiments.

\addtolength{\textheight}{-14.4cm}   




\section*{ACKNOWLEDGMENT}

This work was supported by U.S. National Institutes of Health under the grants number 1R01EB023943-04A1 and partially by JHU internal funds.


\end{document}